\documentclass[letterpaper, 10 pt, conference]{ieeeconf}  

\IEEEoverridecommandlockouts                              

\makeatletter
\let\NAT@parse\undefined
\makeatother
\usepackage[numbers]{natbib}

\usepackage{times}

\usepackage{tikz}
\usepackage{graphicx} 
\usepackage{amsmath} 

\usepackage{makecell}

\usepackage{amssymb}  
\usepackage{amsfonts}
\usepackage{subfigure}
\usepackage{ctable}
\usepackage{makecell}

\usepackage[titlenumbered,linesnumbered,ruled,vlined]{algorithm2e}

\usepackage{fancyvrb}
\usepackage{xcolor}
\usepackage{verbatimbox}

\usepackage{caption}

\usepackage[numbers]{natbib}
\usepackage{multicol}
\usepackage[bookmarks=true]{hyperref}

\usepackage{amsthm}
\theoremstyle{definition}

\usepackage{irap_acronyms}
\usepackage{irap_math}
\usepackage{irap_SIunits}
\usepackage{irap_misc}

\usepackage{soul,color}
\usepackage{verbatim} 
\usepackage{multirow}

\usepackage{lipsum}
\usepackage{xcolor}

\let\oldnl\nl
\newcommand{\nonl}{\renewcommand{\nl}{\let\nl\oldnl}}

\begin{document}

\title{\LARGE \bf
  \text{LT-mapper}: A Modular Framework for LiDAR-based Lifelong Mapping
}

\author{ Giseop Kim${}^{1}$ and Ayoung Kim${}^{1*}$
    \thanks{
      $^{1}$G. Kim is with the Department of Civil and Environmental Engineering, KAIST, Daejeon, S. Korea {\tt\small paulgkim@kaist.ac.kr}
      \newline
      \indent $^{1}$A. Kim is with Department of Civil and Environmental Engineering, KAIST, Daejeon, S. Korea {\tt\small ayoungk@kaist.ac.kr} 
    }%
}

\maketitle
\begin{abstract}



Long-term 3D map management is a fundamental capability required by a robot to reliably navigate in the non-stationary real-world. This paper develops open-source, modular, and readily available LiDAR-based lifelong mapping for urban sites. This is achieved by dividing the problem into successive subproblems: \ac{MSS}, high/low dynamic change detection, and positive/negative change management. The proposed method leverages \ac{MSS} and handles potential trajectory error; thus, good initial alignment is not required for change detection. Our change management scheme preserves efficacy in both memory and computation costs, providing automatic object segregation from a large-scale point cloud map. We verify the framework's reliability and applicability even under permanent year-level variation, through extensive real-world experiments with multiple temporal gaps (from day to year).


\end{abstract}

\IEEEpeerreviewmaketitle

\acresetall

\section{Introduction}
\label{sec:intro}

\begin{figure}[!t]
  \centering
  \captionsetup{font=footnotesize}

  \includegraphics[width=0.99\columnwidth, trim = 0 0 0 0, clip]{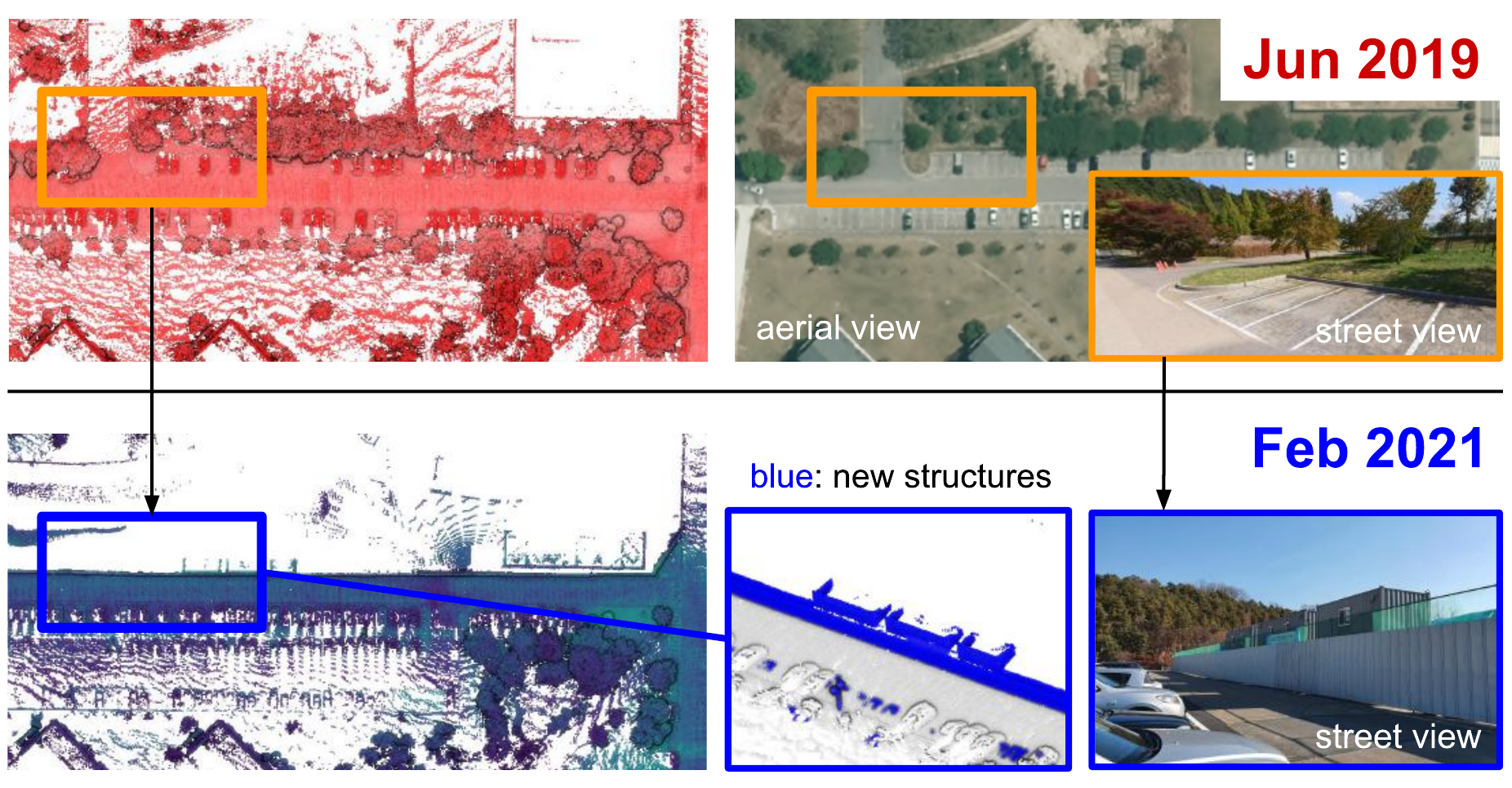}
  \caption
  {
    An example of permanent structural changes over $\sim$ 1.5-year temporal gap. (Top) \texttt{KAIST 01} of \texttt{MulRan} dataset \cite{kim2020mulran} (June 2019). (Bottom) \texttt{KAIST 04}, recently released in extended sequences (February 2021). A construction wall appeared over time; the previously existing parking spaces and trees disappeared. \textbf{LT-mapper} can accurately register the temporally disjointed maps and detect pointwise changes (e.g., blue points in the bottom blue box).
  }
  \label{fig:cover}
  \vspace{-5mm}
\end{figure}

During long-term mapping {using \ac{LiDAR}} sensor, we encounter changes in an environment as in \figref{fig:cover}. The perceived snapshot of the environment contains both ephemeral and persistent objects that may change over time. To handle this change properly, long-term mapping must solve for autonomous map maintenance \cite{pomerleau2014long} by detecting, updating, and managing the environmental changes accordingly. In doing so, the challenges in scalability, potential misalignment error, and map storage efficiency should be addressed and resolved toward lifelong map maintenance.



\textit{1) Integration to multi-session SLAM for scalability:} Some studies regarded change detection as a post-process of comparing multiple pre-built maps associated with temporally distant and independent sessions. As reported in \cite{kim2010multiple}, alignment of multiple sessions in a global coordinate may severely limit scalability. Following their philosophy, in this work, we integrate \ac{MSS} and align sessions with anchor nodes \cite{kim2010multiple} to perform change detection in a large-scale urban environment beyond a small-sized room \cite{ambrucs2014meta}. Our framework consists of a LiDAR-based multi-session 3D \ac{SLAM} module, named \textbf{LT-SLAM}. 


\begin{figure*}[!t]

  \captionsetup{font=footnotesize}

  \centering
  \resizebox{.86\textwidth}{!}{
  \centering
  \subfigure[ { Inconsistent ground-truths over sessions} ]{%
  \includegraphics[width=0.305\textwidth, trim = -40 0 0 0, clip]{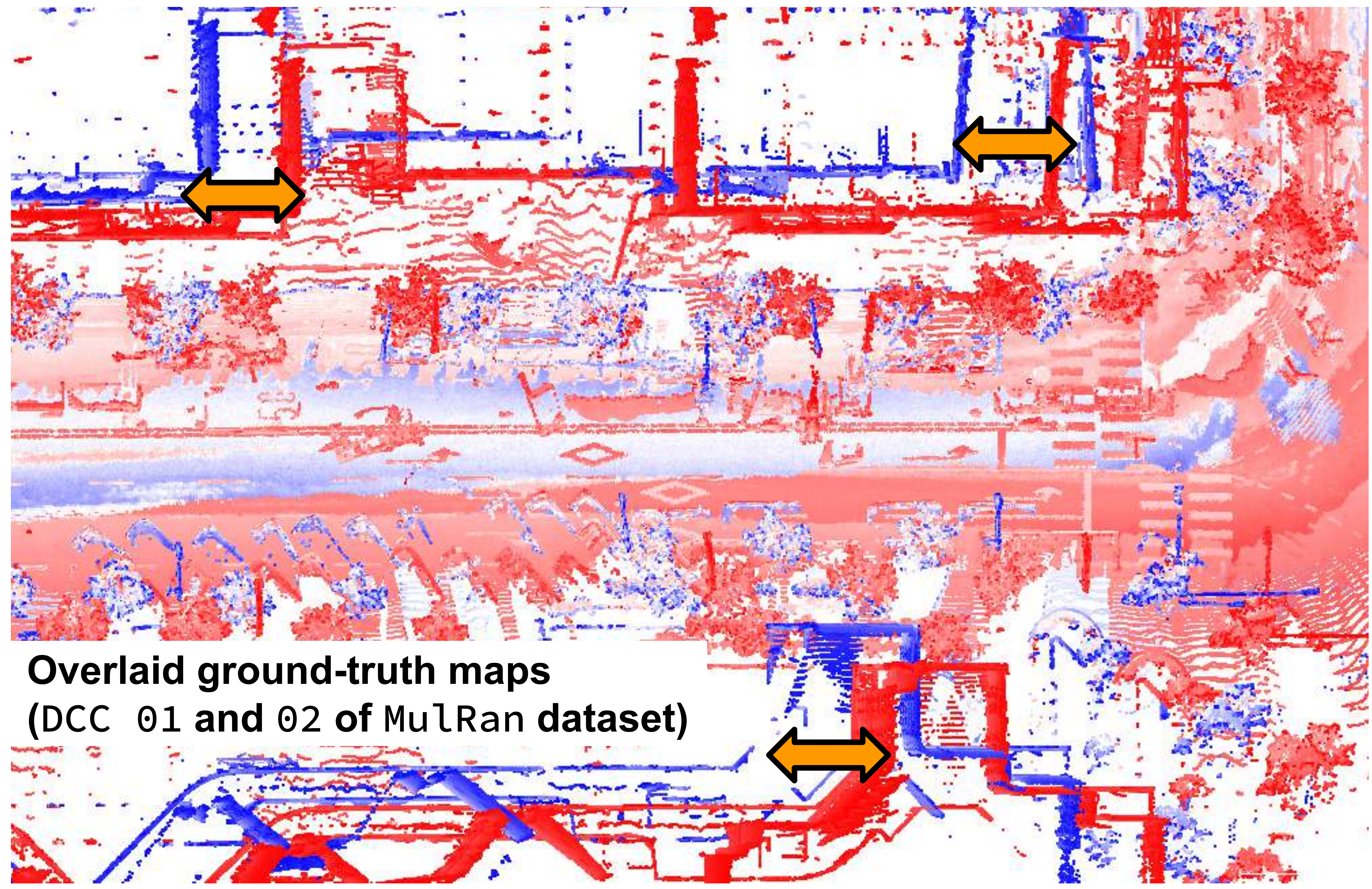}
      \label{fig:ltmapperneeds1}
  } \ \ 
  \subfigure[ { Noisy points from moving objects} ]{%
  \includegraphics[width=0.295\textwidth, trim = 0 0 40 0, clip]{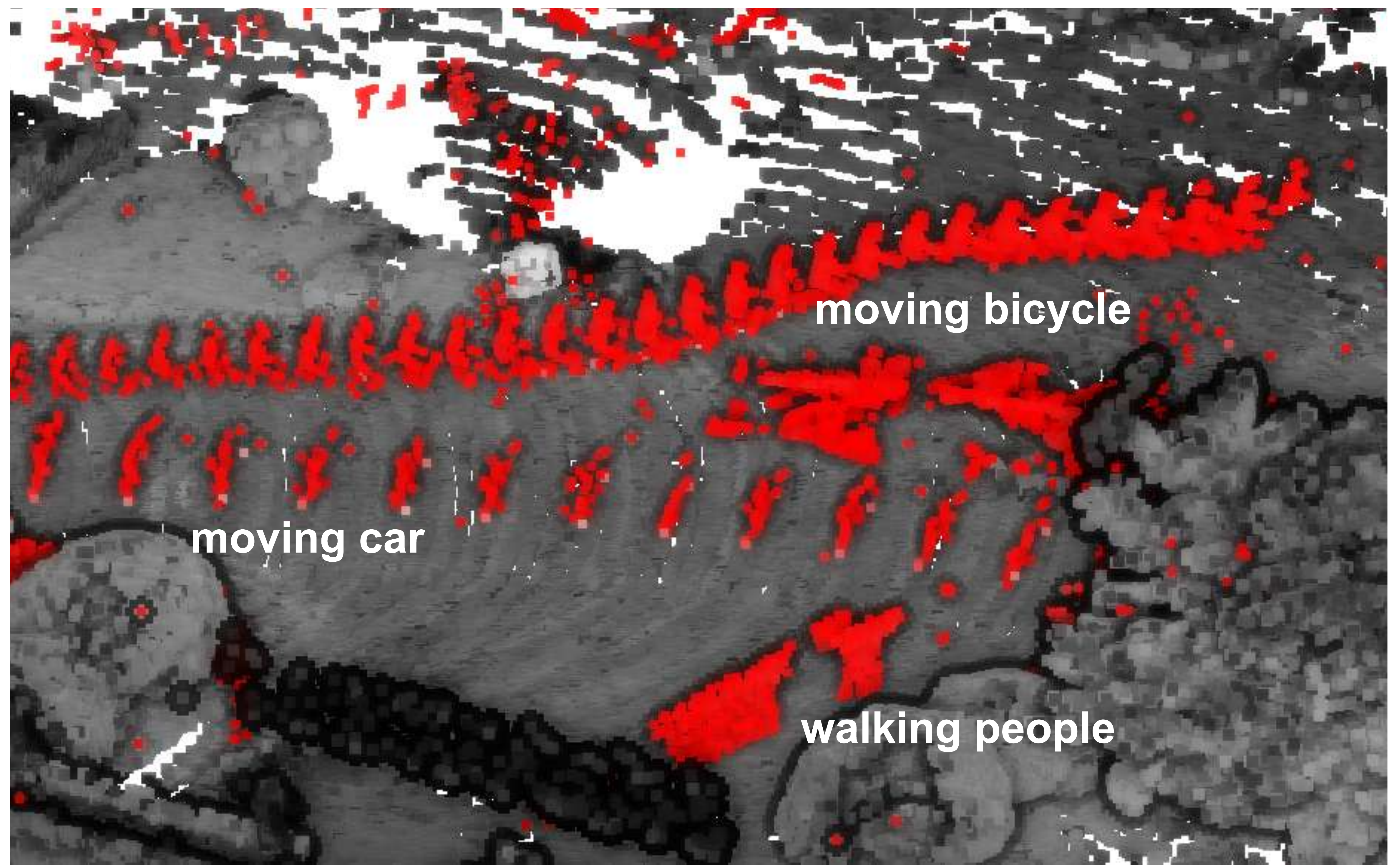}
      \label{fig:ltmapperneeds2}
  } \ \ 
  \subfigure[ { Disappearing objects } ]{%
  \includegraphics[width=0.28\textwidth, trim = 20 40 160 0, clip]{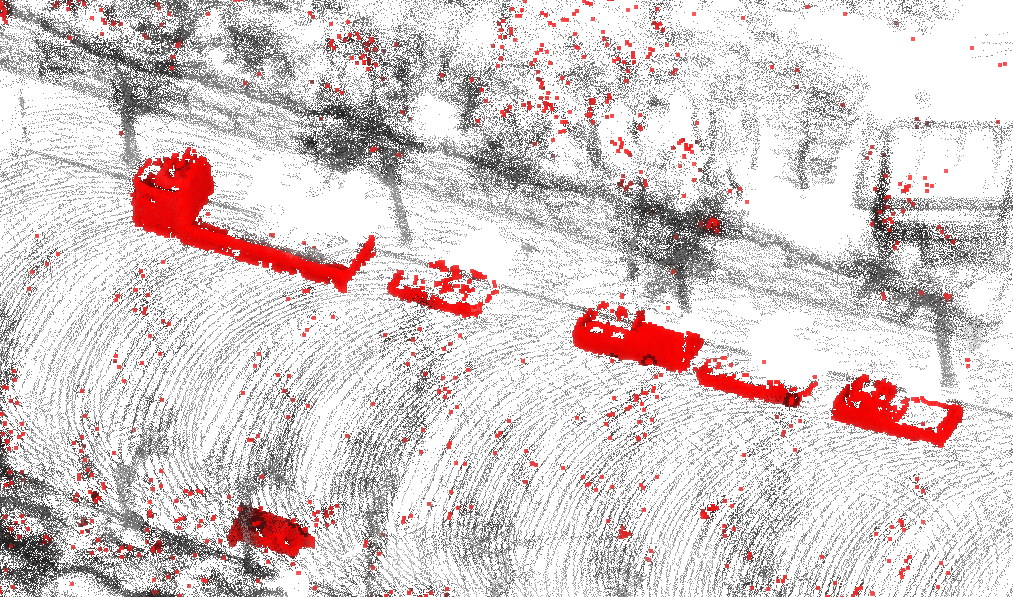}
      \label{fig:ltmapperneeds3}
  }
  }
  \vspace{-2mm}
  \caption
  {
    Three challenges for LiDAR-based lifelong mapping in urban sites. (a) Overlaid \ac{GT} maps of \texttt{MulRan} \cite{kim2020mulran} dataset \texttt{DCC 01} and \texttt{02}. Even with highly accurate sensors (e.g., RTK-GPS), \ac{GT} maps may not be globally consistent along the temporal axis. (b) Moving objects (red dots) linger on a map as ghost points. (c) Structures or objects may disappear (red dots) or appear anew at a different session. These changes should be updated properly. 
  }
  \label{fig:ltmapperneeds}

  \vspace{-7mm}

\end{figure*}



\textit{2) Change detection under SLAM error:} Change detection between two maps would be trivial if maps were perfectly aligned. Early works \cite{walcott2012dynamic, ambrucs2014meta, wellhausen2017reliable, schauer2018peopleremover} in map change detection relied on the strong assumption of globally well-aligned maps with no error and avoided handling this ambiguity issue. Unfortunately, trajectory error inevitably occur in reality. We reconcile this potential misalignment during our change detection, and enable the proposed method to handle potential alignment error robustly. To deal with the ambiguity, we propose a scan-to-map scheme with projective visibility, using range images of multiple window sizes named \textbf{LT-removert}. By extending an intra-session change detection method \cite{gskim-2020-iros}, the \text{LT-removert} includes both intra-/inter-session change detection, thereby further segregating high and low-dynamic objects \cite{walcott2012dynamic} from the change.

\textit{3) Compact place management:} In addition to change detection, we present and prove a concept of change composition. Once the change is detected, the decision for map maintenance should be followed to determine what to include or exclude. Using this feature, ours not only maintains an up-to-date map such as existing works \cite{pomerleau2014long, ambrucs2014meta}, but also extracts stable structures with higher \textit{placeness}; thereby, we construct a reliable 3D map with authentically meaningful structures for other missions, such as cross-modal localization \cite{kim2018stereo} and long-term localization \cite{kim2018scan}. This final module, named \textbf{LT-map}, manages the changes and enables a central map to evolve in a place-wise manner.

 In sum, we propose a novel modular framework for LiDAR-based lifelong mapping, named \textbf{LT-mapper}. Each module in the framework can run separately via file-based in/out protocol. Unified and modular lifelong mapping has barely been made for 3D \ac{LiDAR}, unlike recently (but partially) delivered visual-based methods \cite{Stent-RSS-16, labbe2019rtab, schneider2018maplab, orbatlas19, orbslam3}. To the best of our knowledge, \text{LT-mapper} is the first open modular framework that supports \ac{LiDAR}-based lifelong mapping in complex urban sites. The proposed has the following contributions:

\begin{itemize}
    \item We integrate \ac{MSS} with change detection and handle sessions resiliently via anchor node. The submodule \textbf{LT-SLAM} can stitch multiple sessions in a shared frame using only LiDAR.

    \item The submodule \textbf{LT-removert} overcomes alignment ambiguity between sessions with remove-then-revert algorithm along spatial and temporal axes.

    \item The submodule \textbf{LT-map} can produce both an up-to-date map (\textit{live map}) and a persistent map (\textit{meta map}) efficiently, while storing changes as a \textit{delta map}. By exploiting delta maps, restoration and change detection become memory and computation cost-effective.

    \item The aforementioned modules are packaged within a single framework, and it is publicly released\footnote{The code is available at https://github.com/gisbi-kim/lt-mapper.} with readily available console-based commands. Also, we provide real-world experiments with multiple temporal gaps (day to year).
\end{itemize}

\vspace{-1mm}
\section{Related Works}
\label{sec:related}



\subsubsection{Multi-session SLAM}

In \cite{pomerleau2014long, ambrucs2014meta}, a query scan is assumed to be well localized within the map. However, in the real-world outdoor environment, \ac{SLAM} error exist and registration between scans may be vulnerable, failing even with small and partial structural variance. Thus, as claimed in \cite{kim2010multiple, orbatlas19}, jointly smoothing the multiple sessions can improve query-to-map localization performance despite potential motion drift \cite{walcott2012dynamic}. 





\subsubsection{3D Change Detection}

Given well-aligned maps, a set difference operation can be conducted via extracting map-to-map complements \cite{ambrucs2014meta, wellhausen2017reliable}. Otherwise, visibility-based scan-to-map discrepancy comparison \cite{pomerleau2014long, schauer2018peopleremover, gskim-2020-iros, palazzolo2018fast} has been a popular choice, because of the small covisible volume and inherent localization errors. Removert \cite{gskim-2020-iros} leveraged range images of multiple window sizes. However, it was restricted to a single session and has not treated high and low dynamic points separately. 


\subsubsection{Lifelong Map Management}

Lifelong map management should consider two factors: \textit{1)} which entity (representation), \textit{2)} how to be updated (update unit)? 

\noindent \textbf{Representation}. The long-term map representation varies from traditional occupancy grid maps \cite{tipaldi2013lifelong, sun2018recurrent, banerjee2019lifelong} to frequency domain representation \cite{krajnik2017fremen}. For change detection in 3D environments, dealing with a direct raw 3D point cloud may be preferred over the occupancy map-based ones. 

\noindent \textbf{Update Unit}. With respect to the atomic map update unit, we manage changes at a keyframe level. This contributes to systems scalability, without being restricted to a fixed global frame \cite{pomerleau2014long} or room level \cite{ambrucs2014meta}.






\subsubsection{Modular Design of Lifelong Mapper}

The abovementioned exiting modules have been developed individually, whereas a unified system has hardly been discovered for LiDAR. DPG-SLAM \cite{walcott2012dynamic} combined the full modules but was constrained in SE(2) space and lack of 3D change detection. \cite{pomerleau2014long} was also equipped with entire modules, except for \ac{MSS}.


\vspace{-1mm}
\section{Overview}
\label{sec:overview}

\begin{figure}[!b]

  \vspace{-5mm}

  \centering
  \captionsetup{font=footnotesize}

  \includegraphics[width=0.98\columnwidth]{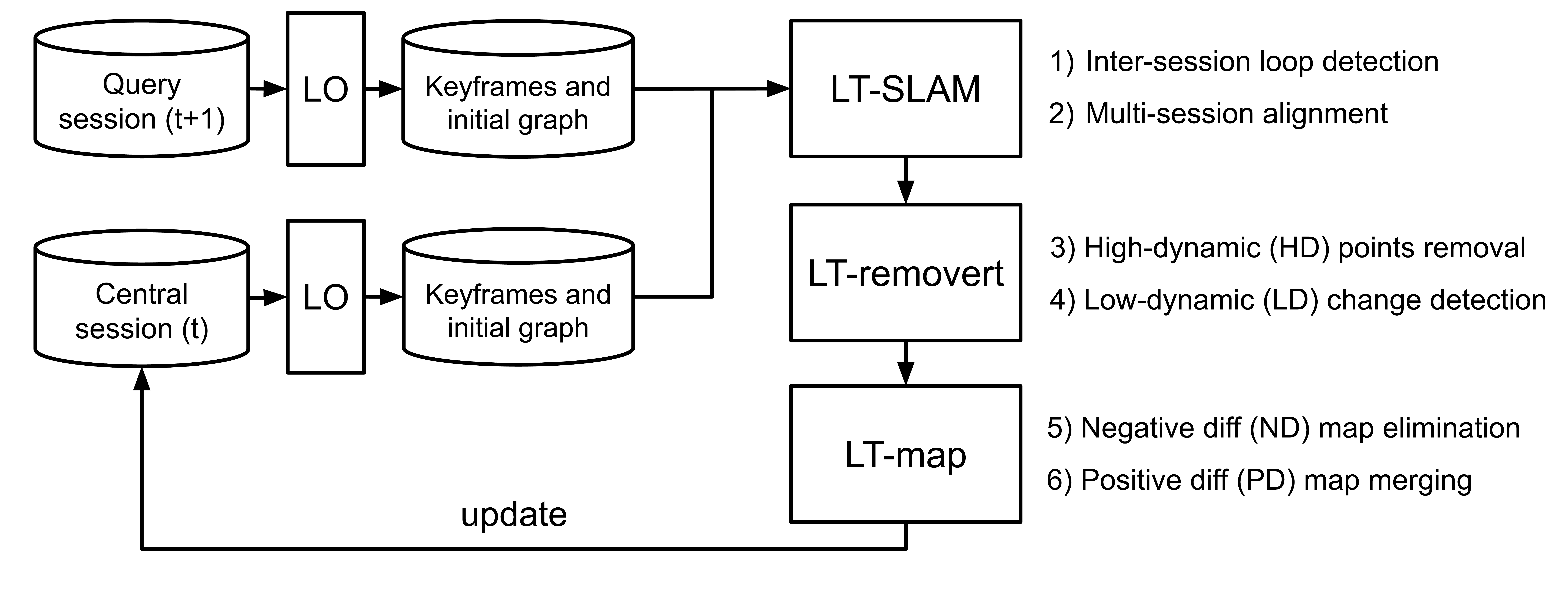}
  \caption
  {
    A modular pipeline of \textbf{LT-mapper} system. The framework is composed of three modules: \textbf{LT-SLAM}, \textbf{LT-removert}, and \textbf{LT-map}.
  }
  \label{fig:pipeline}
\end{figure}

LT-mapper is fully modular and supports the three aforementioned functionalities. The overall pipeline is composed of three modules (\figref{fig:pipeline}), which run sequentially and independently. Unlike existing LiDAR-based change detection \cite{ding2020lidar} equipped with expensive localization suites, our system requires only a single LiDAR sensor (optionally IMU for odometry at the initial pose-graph construction).

Accurate alignment between temporarily disconnected sessions is elusive in real-world outdoor environments, as can be seen in \figref{fig:ltmapperneeds1}. In \text{LT-SLAM} module, we utilize multi-session \ac{SLAM} that jointly optimizes multiple sessions accompanied with robust inter-session loop detection from a LiDAR-based global localizer. In this module, a query measurement is registered to the existing central map.

We also need to consider the measurement volatility. For example, in \figref{fig:ltmapperneeds2}, a contructed point cloud map may be noisy, due to surrounding moving objects (red dots) even with the accurate odometry. These volatile objects contribute less to a place's distinctiveness than stationary points. Thus, these \textit{\ac{HD}} points should be pre-removed before the between-session-differences calculation in \text{LT-removert} module.

After aligning a query and a central session and removing the HD points, we detect changes by applying set difference operation between query measurements and a central map, as in \figref{fig:ltmapperneeds3}. We call the change \textit{\ac{LD}}, and it is further divided into two classes: newly appeared points (\textit{\ac{PD}}) and disappeared points (\textit{\ac{ND}}).




\vspace{-1mm}
\par\noindent\rule{\columnwidth}{0.4pt}
\begin{Verbatim}[fontsize=\footnotesize, commandchars=\\\(\)]
  \color(blue)# Read single-session graphs and their keyframes
  ./ltslam       # with params_ltslam.yaml    
  \color(blue)# Save aligned graphs 

  \color(blue)# Read the aligned graphs and keyframes' submaps
  ./ltremovert   # with params_ltremovert.yaml  
  \color(blue)# Save PD, ND keyframe scans

  \color(blue)# Read the PD, ND keyframe' submaps
  ./ltmap        # with params_ltmap.yaml   
  \color(blue)# Save updated submaps [and merged maps for viz]
\end{Verbatim}
\vspace{-2mm}
\par\noindent\rule{\columnwidth}{0.4pt}

\vspace{-1mm}
\section{LT-mapper}
\label{sec:ltmapperall}

In this section, we give details of the three modules of \text{LT-mapper}. We define a session $\mathcal{S}$ as
\begin{equation}
    \mathcal{S} := ( \mathcal{G}, \{ (\mathcal{P}_i, d_i) \}_{i=1, ..., n} ) \ ,
    \label{eq:sess}
\end{equation}
where $\mathcal{G}$ is a pose-graph text file (e.g., \texttt{.g2o} format) containing a set of pose nodes' indexes and initial values, odometry edges, and optionally putative intra-session loop edges. This initial pose-graph can be constructed by using any existing LiDAR (-inertial) odometry algorithms \cite{shan2018lego,cho2020unsupervised, li2020dmlo, shan2020lio, yokozukalitamin}. We allow potential navigational drifts and overcome the intra-session drifts via multi-session pose-graph optimization. The $(\mathcal{P}_i, d_i)$ are a 3D point cloud $\mathcal{P}$ and the its global descriptor $d$ (e.g., \cite{he2016m2dp, kim2018scan,angelina2018pointnetvlad,Chen2019OverlapNetLC,xu2020disco}) for the $i^\text{th}$ keyframe. We assign an equidistant sampled keyframes and $n$ is the number of total keyframes.

\subsection{LT-SLAM: A Multi-session SLAM Engine}
\label{sec:ltslam}


We denote the existing session $\mathcal{S}_{t_c}$ at time $t_c$ as \textit{central} ($C$), and the newly obtained session $\mathcal{S}_{t_q}$ at time $t_q \ > t_c$ as \textit{query} ($Q$). Given a pair of the central and query sessions, \text{LT-SLAM} aligns the two sessions. 

\begin{figure}[!t]
  \centering
  \captionsetup{font=footnotesize}

    \centering
    \subfigure[ { \texttt{KAIST} sequences of \texttt{MulRan} dataset } ]{%
    \includegraphics[width=0.9\columnwidth, trim = 0 0 -10 0, clip]{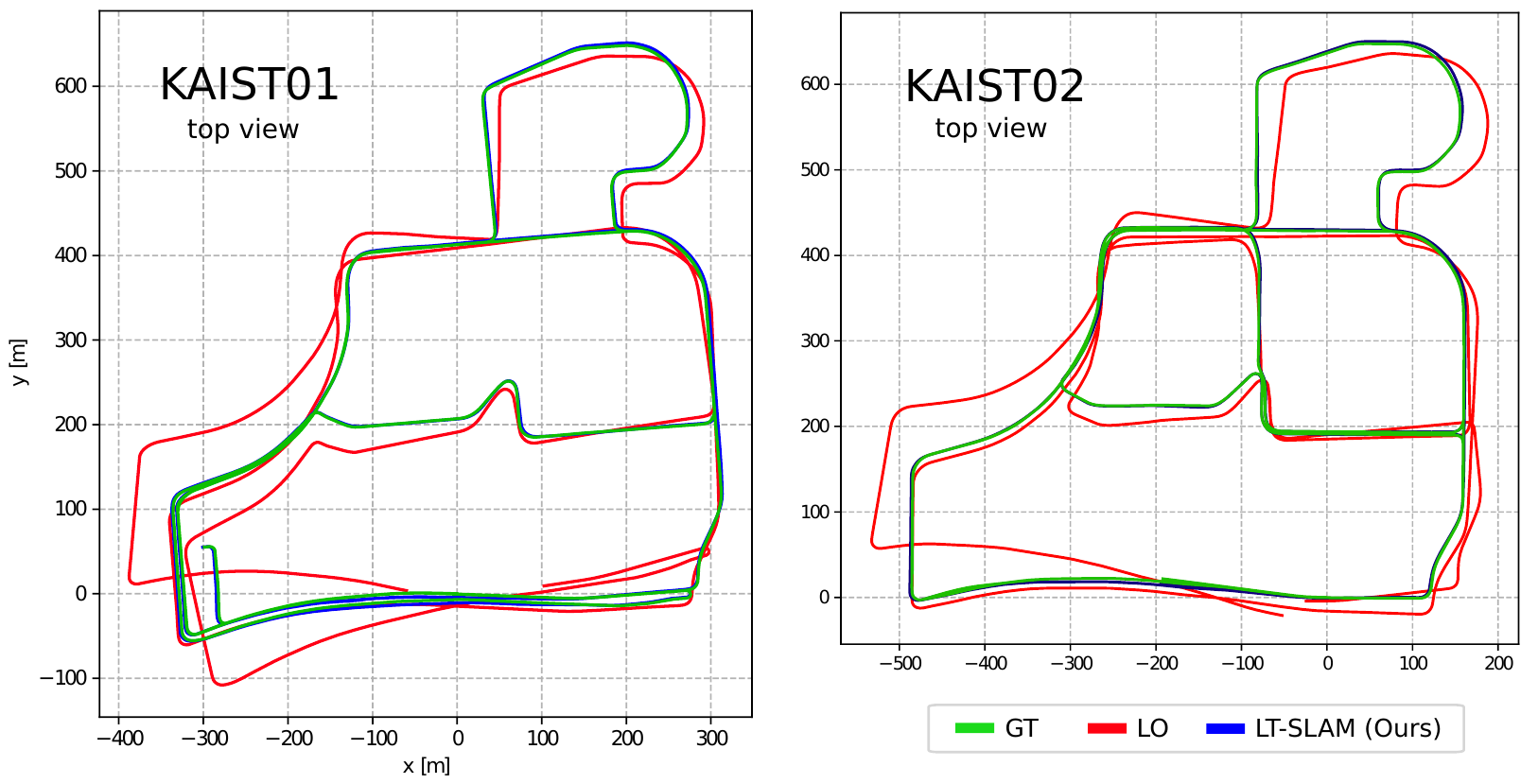}
        \label{fig:ltslamqual1}
    } \\ 
    \subfigure[ { \texttt{LT-ParkingLot} dataset } ]{%
    \includegraphics[width=0.95\columnwidth, trim = -30 0 20 0, clip]{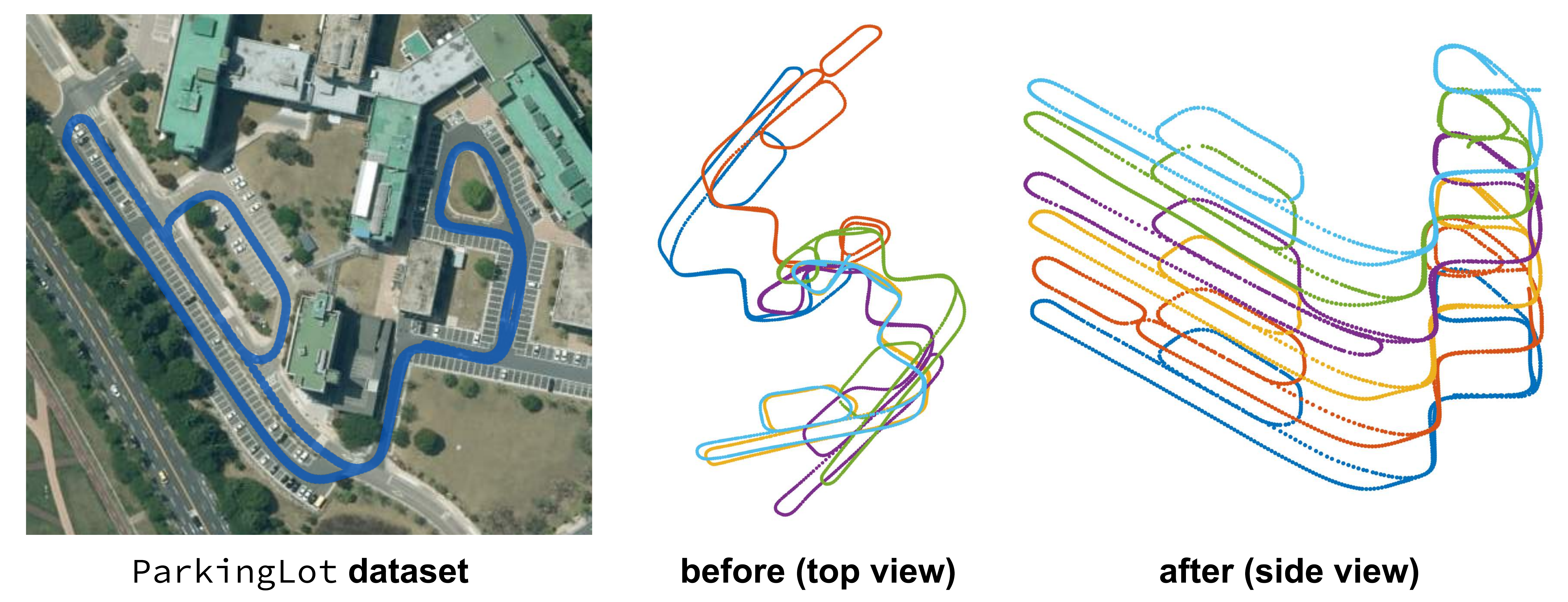}
        \label{fig:ltslamqual2}
    }

  \caption
  {
    \subref{fig:ltslamqual1} The baseline \texttt{LO} (red) is obtained by LIO-SAM's \cite{shan2020lio} odometry without loop closures. Scan Context-based inter-session loops successfully mitigated each session's internal drifts. \subref{fig:ltslamqual2} The left is an aerial view of \texttt{LT-ParkingLot} dataset. The middle plot shows different sessions' trajectories acquired at different dates. While covering the same area, trajectories were not aligned, as they occupy their own coordinates. The right plot shows that \text{LT-SLAM} simultaneously estimates between-session offsets while reducing the drifts. Thus, multiple sessions can be aligned in a shared world coordinate for change detection. The height differences are imposed for clear visualization.
  }
  \label{fig:ltslamqual}

  \vspace{-5mm}

\end{figure}


\begin{figure*}[!t]
  \centering
  \captionsetup{font=footnotesize}

  \includegraphics[width=0.99\textwidth]{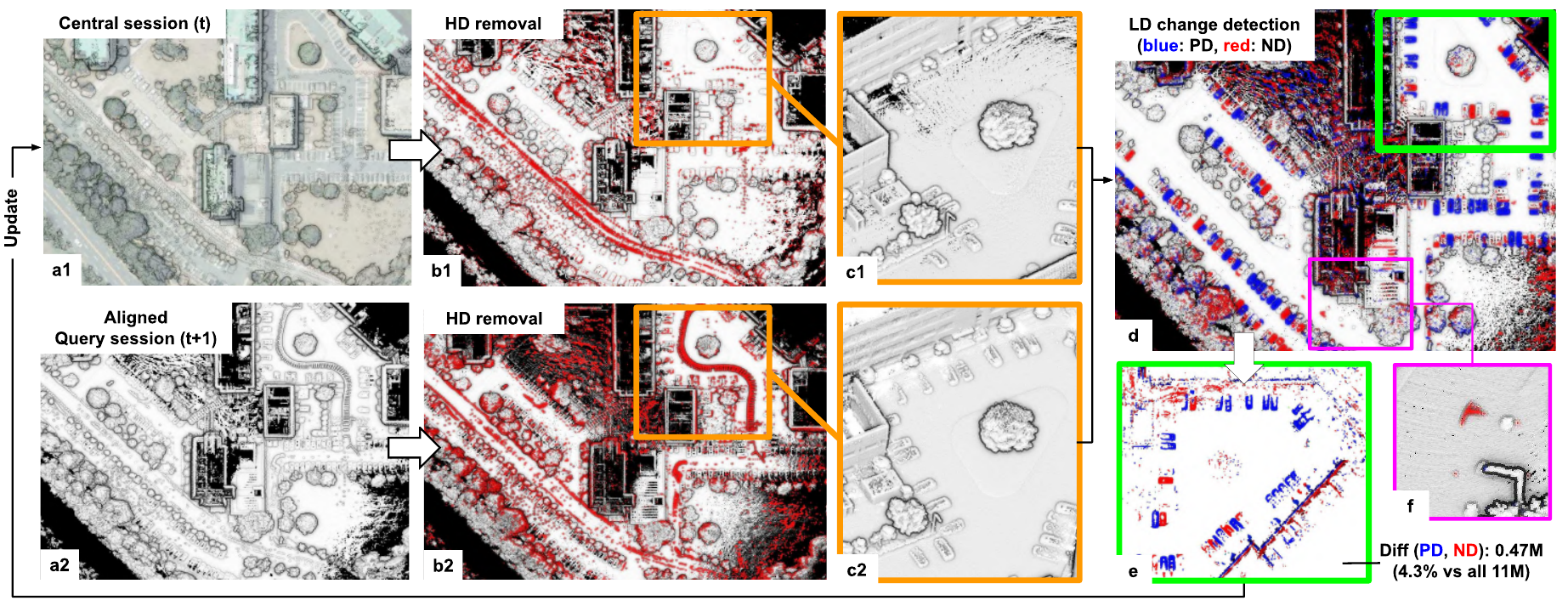}
  \caption
  {
    A visualization of \textbf{LT-removert} pipeline. (a) \text{LT-removert} receives aligned central and query maps from \text{LT-SLAM}. (b, c) cleaned maps with \ac{HD} points removed. (d, e) \ac{LD} change detection (i.e., \ac{PD} and \ac{ND} segmentation). (f) deletion of unremoved \ac{HD} points via multi-sessions.
  }
  \label{fig:ltemovertvisualpipeline}

  \vspace{-5mm}

\end{figure*}


The incoming sessions' pose-graphs preserve their own coordinates and \text{LT-SLAM} utilizes the anchor node-based inter-session loop factors \cite{kim2010multiple, mcdonald2013real, ozog2016long}. As \citeauthor{kim2010multiple} \cite{kim2010multiple} reported, the anchor node can successfully estimate a between-session offset, resolving their intra-session drifts.
The anchor node-based loop factor for a relative pose measurement $z$ is
\begin{equation}
    \begin{aligned}
        & \phi( \textbf{x}_{C, i}, \textbf{x}_{Q, j}, \Delta_{C}, \Delta_{Q}) \\
        & \propto \text{exp} \left( -\frac{1}{2} \| \left( (\Delta_{C}\oplus\textbf{x}_{C, i})\ominus(\Delta_{Q}\oplus\textbf{x}_{Q, j})\right) - z \|^2_{\Sigma_z} \right) \ ,
    \end{aligned}
    \label{eq:anchorfactor}
\end{equation}
where $\textbf{x}$ means a SE(3) pose, $i$ and $j$ are pose variable indexes. $\oplus$ and $\ominus$ are the SE(3) pose composition operators \cite{smith1990estimating}. $\Delta$ indicates an anchor node, which is also a SE(3) pose variable. The central session's anchor node $\Delta_{C}$ has very small covariance while the query's $\Delta_{Q}$ has a very large value.

We need to identify a loop-closure candidate $(i, j)$ between sessions $C$ and $Q$. For robust \text{inter-session loop detection}, we adopt \ac{SC} \cite{kim2018scan} due to their long-term global localization capability and light computation cost. After the inter-session loop is detected, a 6D relative constraint between two keyframes is calculated via \ac{ICP} using their submap point clouds $\mathcal{P}_{C, i}$ and $\mathcal{P}_{Q, j}$. We only accept loops with acceptably low ICP's fitness scores, and use the score for an adaptive covariance $\Sigma_{z}$ in \eqref{eq:anchorfactor}. We also use robust back-end (e.g., \cite{agarwal2013robust, mangelson2018pairwise}) for all inter-session loop factors for safe optimization under inevitable false loop detections. Given the initially aligned sessions using SC-loops, we further refine the graph using radius search loop detection (i.e., based on pose proximity) for non-\ac{SC}-detected keyframes to finely stitch the sessions.

Finally, each session's trajectory is optimized within their own coordinates (denoted ${}^{C}\mathcal{G}_{C}^{*}$ and ${}^{Q}\mathcal{G}_{Q}^{*}$) as in \figref{fig:ltslamqual1}. The optimized maps are then represented in a shared world coordinate $W$ to be consumed by \text{LT-removert} introduced in \secref{sec:ltremovert}. To do so, \text{LT-SLAM} returns pose-graphs ${}^{W}\mathcal{G}_{C}^{*}$ and ${}^{W}\mathcal{G}_{Q}^{*}$ by applying the below transforms for each pose $\textbf{x}$ in a graph:
\begin{equation}
    \begin{aligned}
        {}^{W}\textbf{x}_{C}^{*} = \Delta_{C}^{*} \oplus {}^{C}\textbf{x}_{C}^{*} \text{ and }{}^{W}\textbf{x}_{Q}^{*} = \Delta_{Q}^{*} \oplus {}^{Q}\textbf{x}_{Q}^{*} \ .
    \end{aligned}
    \label{eq:centraltransform}
\end{equation}
The right in \figref{fig:ltslamqual2} shows the aligned multiple sessions sharing the same coordinates.

\subsection{LT-removert: Two-session Change Detection}
\label{sec:ltremovert}

As mentioned in \secref{sec:overview}, the dynamic points are classified into \ac{HD} and \ac{LD}. 
In our second module, \text{LT-removert}, we first remove \ac{HD} without erasing \ac{LD} points. We denote $\text{LD}_{C}^{Q}$ means 3D points that are low dynamic changes detected at a place (keyframe) between the session $C$ (from) and $Q$ (to).




\subsubsection{High Dynamic (HD) Points Removal}
\label{sec:ltremovert1}

We choose Removert \cite{gskim-2020-iros} for our \ac{HD} points removal engine. Using range image-based discrepancy, Removert utilizes different sizes of windows to alleviate the pose ambiguity. For example, \figref{fig:ltemovertvisualpipeline} (b) and (c) show before and after of applying Removert.



\subsubsection{Low Dynamic (LD) Change Detection}
\label{sec:ltremovert2}

Once two sessions are aligned and \ac{HD} points within them were removed, we compare query and central sessions to parse \ac{LD} points. To do so, we construct a kd-tree for the target map and test whether a source map's point has $k$ target map points within a threshold $r$\unit{}{m} (if not, the point is LD). Then, the \ac{ND} and \ac{PD} points are parsed. 



\subsubsection{Weak ND Preservation (Handling Occlusions)}
\label{sec:ltremovert3}

\begin{figure}[!t]
  \centering
  \captionsetup{font=footnotesize}

  \subfigure[ Two snapshots ]{%
  \includegraphics[width=0.3\columnwidth, trim = 0 -35 0 0, clip]{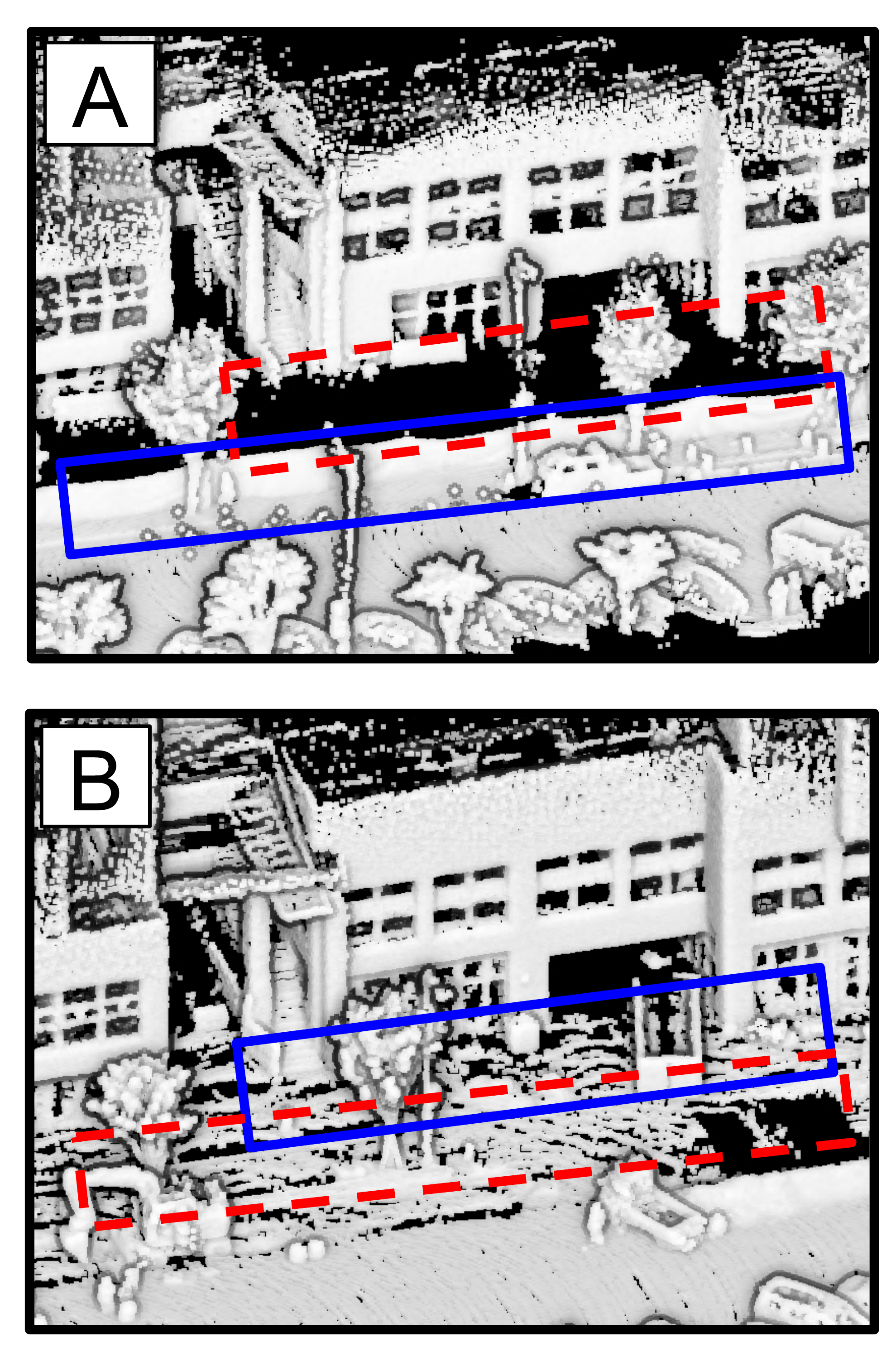}
      \label{fig:ldcases1}
  } \
  \subfigure[ Two possible change scenarios ]{%
  \includegraphics[width=0.62\columnwidth, trim = 0 0 0 0, clip]{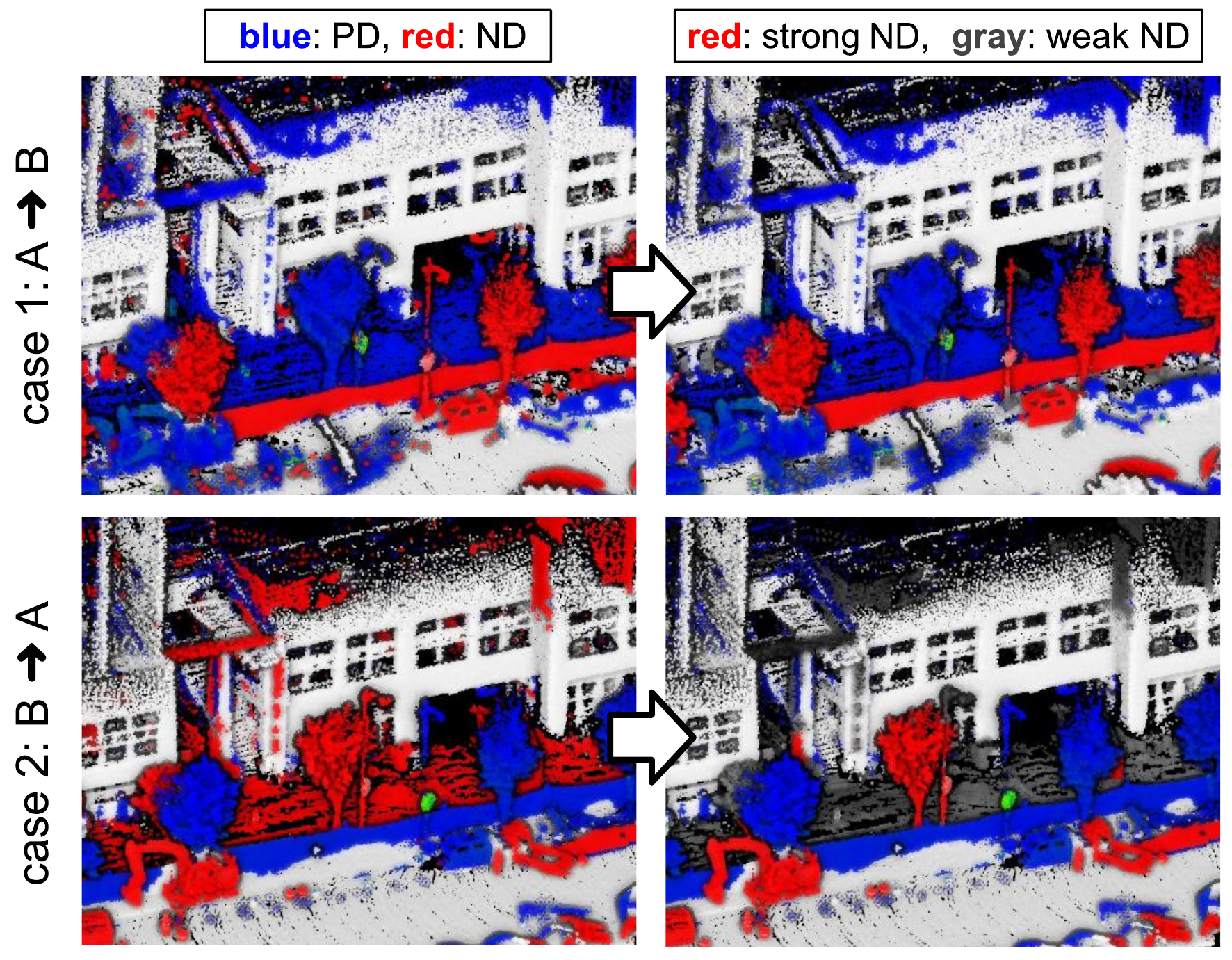}
      \label{fig:ldcases2}
  }

  \caption
  {
    \subref{fig:ldcases1} The exemplars (\texttt{A}) and (\texttt{B}) from \texttt{DCC 01} and \texttt{02} of \texttt{MulRan} dataset. (\texttt{A}) includes occlusion where no points exist (red dotted box), due to the construction wall (blue-lined box). The wall disappears later, revealing the ground points in (\texttt{B}). \subref{fig:ldcases2} Selection of central session as \texttt{A} and query session as \texttt{B} will be \texttt{Case 1}; the reversed case is \texttt{Case 2}. The left column is the naive ND and PD points from the set difference operation. The right column shows accepted ND and PD points after weak ND point validation. In \texttt{Case 1}, red ND points were mostly removed while adding entire blue PD points. In \texttt{Case 2}, some ground points are occluded and falsely marked as ND in the left column. In the right column, only strong ND (red) points are removed, and weak ND points are reverted (i.e., ground points marked with gray).
  }
  \label{fig:ldcases}

  \vspace{-5mm}

\end{figure}

Another critical factor to consider is occlusion as argued in \cite{ambrucs2014meta}. In \figref{fig:ldcases}, an example is given to show the effect of occlusion in determining valid \ac{PD} and \ac{ND}. Naturally, the central session \texttt{A} will be compared against the query session \texttt{B} (\texttt{case 1}). However, let us consider the reversed case (\texttt{case 2}) when \texttt{B} occurs prior to \texttt{A}. In this case, some ground points were occluded by walls and became \ac{ND} points. However, these ground points should not be removed. We name them as \textit{weak ND} and examine further segregation to avoid falsely removing occluded static points.



 For this step, we again employ Removert but with modification. Unlike the original Removert, which removes near map points, the modified Removert removes further points in the raw ND map and reverts them to the static map. The bottom right in \figref{fig:ldcases} shows the preserved weak ND points (gray) being correctly reverted to the static map.
 
 

\subsubsection{Strong PD for Meta-map Construction}
\label{sec:ltremovert4}

\begin{figure*}[!t]
  \centering
  \captionsetup{font=footnotesize}


  \includegraphics[width=0.93\textwidth, trim = 10 10 20 0, clip]{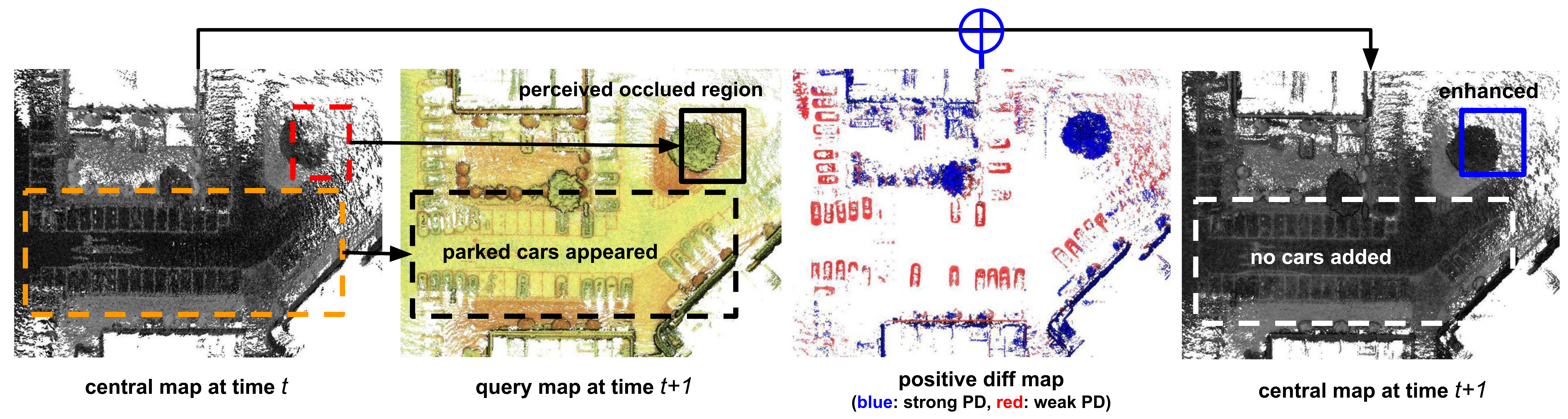}
  \vspace{-2mm}
  \caption
  {
    The sample scene from \texttt{LT-ParkingLot} dataset. Strong PD includes newly captured permanent structures; weak PD tends to be short-term stationary or periodic changes (e.g., parked cars).
  }
  \label{fig:strongpd}
  \vspace{-5mm}
\end{figure*}

We can consider a similar strong/weak classification for PD that is related to whether it retains permanent static structures. We call \text{strong PD} for the points spatially behind. If we only retain strong PD, as in \figref{fig:strongpd}, we can construct a map with maximum volume by carving out the space conservatively. In that sense, we can construct two types of static maps: \textit{meta map} by removing weak PD and \textit{live map} by retaining weak PD. The examples of \textit{meta map} and \textit{live map} are drawn in \figref{fig:metamap}.


\subsection{LT-map: Map Update and Long-term Map Management}
\label{sec:ltmap}


Given the detected LD, \text{LT-map} performs a between-session change update for each keyframe of the central session. The between-session change composition operator $\odot$ is defined as:
\begin{equation}
    \begin{aligned}
        \mathcal{P}_{C} = \tilde{\mathcal{P}}_{C} \odot \text{LD}^{Q}_{C} \triangleq \tilde{\mathcal{P}}_{C} - \textit{ND}(\text{LD}^{Q}_{C}) + \textit{PD}(\text{LD}^{Q}_{C}) \ ,
    \end{aligned}
    \label{eq:changeop}
\end{equation}
where $\tilde{\mathcal{P}}_{C}$ is a keyframe's HD removed point cloud.  The function \textit{ND}$(\cdot)$ and \textit{PD}$(\cdot)$ return ND and PD points near the keyframe and represented in the keyframe's local coordinate. The $-$ and $+$ are set difference and union operation on 3D point space. This \textit{delta map} containing only differences benefits compared to the snapshot-based methods that up/download the whole map. For example, in \figref{fig:ltemovertvisualpipeline}, transmitting the whole new map to a server requires \unit{11}{M} points, whereas only \unit{0.47}{M} points are needed when using \textit{delta maps} (only \unit{4.3}{\%} of the entire map).






\section{Experimental Results}
\label{sec:result}



\subsection{Implementation Detail and Dataset}
\label{sec:expimpl}

\subsubsection{Implementation Detail} Our entire modules are written in C++, and are designed to be readily used with handy commands as in \secref{sec:overview}. \text{LT-SLAM}'s pose-graph optimizer is implemented using iSAM2 \cite{kaess2012isam2} of GTSAM \cite{dellaert2012factor}. We adopted publicly available sources of Scan Context \cite{kim2018scan} and Removert \cite{gskim-2020-iros}. We refer the readers our open source codes\footnote{https://github.com/gisbi-kim/lt-mapper} for the specific parameters of the system. For the initial graph construction to be fed as an input of LT-mapper, we provide keyframe information saver\footnote{https://github.com/gisbi-kim/SC-LIO-SAM} (i.e., pose-graph and global descriptors) as add-ons of existing LiDAR odometry open sources (e.g., LIO-SAM \cite{shan2020lio}). 



\begin{figure}[!b]
  \centering
  \captionsetup{font=footnotesize}

  \vspace{-5mm}

  \includegraphics[width=0.85\columnwidth]{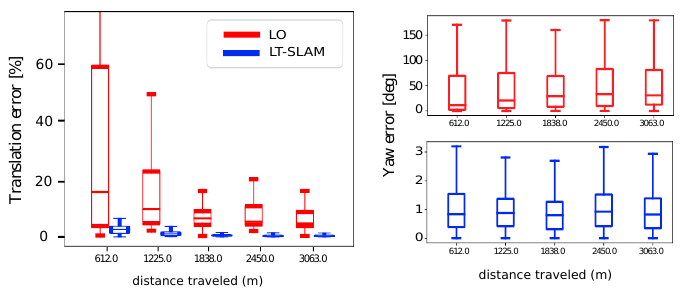}
  \caption
  {
    The quantitative results of LT-SLAM on \texttt{KAIST 01} and \texttt{KAIST 02}. The baseline LO is obtained on \texttt{KAIST 01} by running LIO-SAM \cite{shan2020lio} without loop closures. We note that the error was resolved via \ac{MSS} and inter-session loops despite no initial alignment of \texttt{KAIST 02} being known.
  }
  \label{fig:ltslamquant}


\end{figure}


\subsubsection{Datasets} For the validation, \texttt{MulRan} \cite{kim2020mulran} and our own \texttt{LT-ParkingLot} dataset were selected. Both datasets have multiple sequences and repeated coverage on fixed sites. 

\noindent \textbf{\texttt{MulRan} dataset:} We leveraged this dataset to evaluate the feasibility of our \text{LT-SLAM}. Recently, we have acquired and released an extended sequence for \texttt{KAIST} that is suitable for long-term change detection research. We used the \texttt{KAIST} and \texttt{DCC} sequences to identify long-term changes.

\noindent \textbf{\texttt{LT-ParkingLot} dataset:} A parking lot would be a typical place to witness \ac{LD} changes. We collected six sessions at different times over three days. The sessions' origins are all different and their global alignments are initially unknown as in the middle of \figref{fig:ltslamqual2}.



\subsection{Multi-session Trajectory Alignment}
\label{sec:expltslam}

Both qualitative and quantitative results for \text{LT-SLAM} are shown in \figref{fig:ltslamqual} and \figref{fig:ltslamquant}. We used the RPG trajectory evaluation tool \cite{{Zhang18iros}}. The intra-session translation and rotation (particularly yaw) errors are noticeably reduced via the inter-session anchor node-based loops. Two sessions with different origins successfully suppressed each other's drifts.




\begin{figure}[!b]
  \centering
  \captionsetup{font=footnotesize}

  \vspace{-5mm}

  \includegraphics[width=0.9\columnwidth, trim = 0 0 0 0, clip]{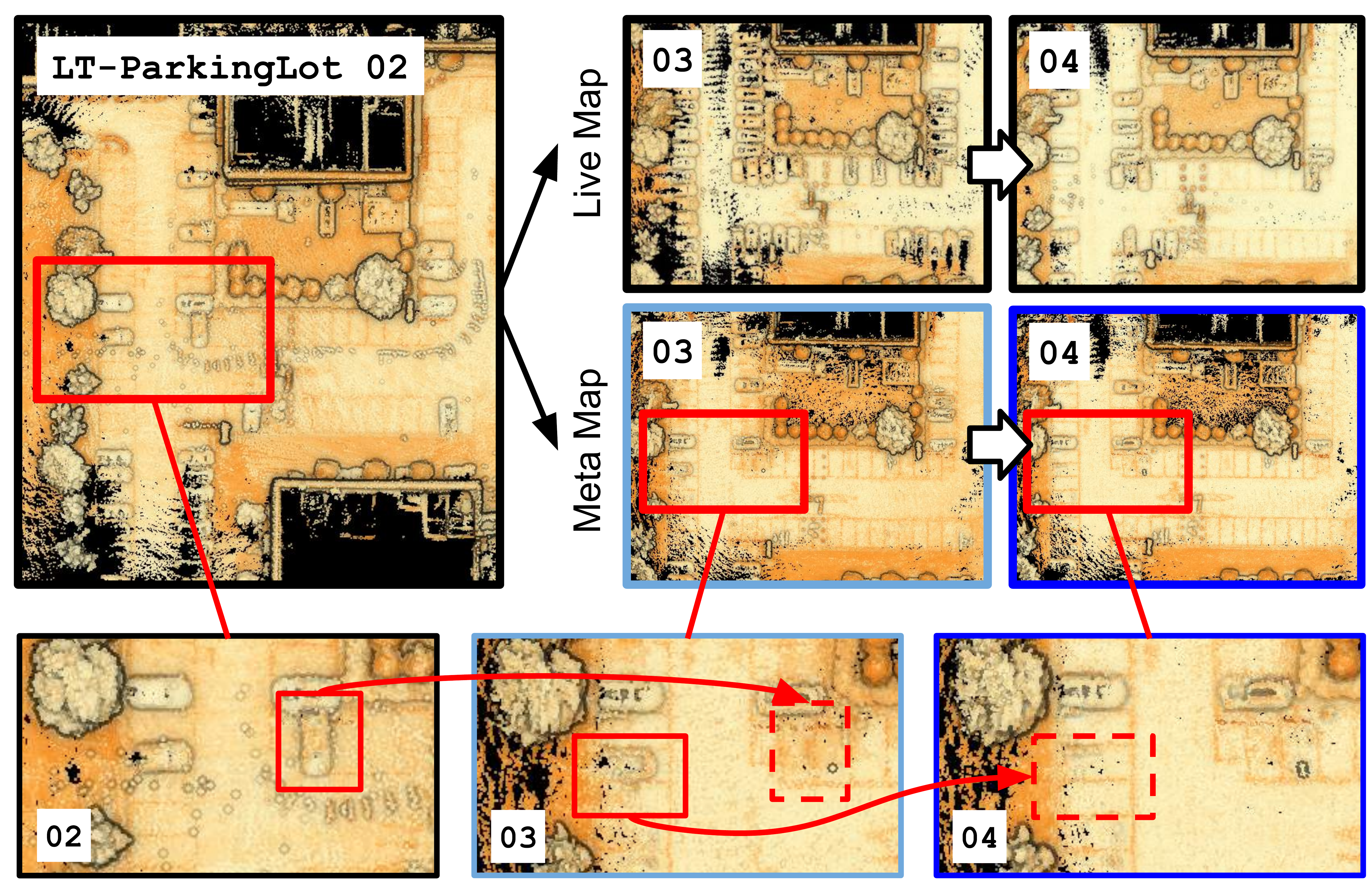}
  \caption
  {
    \text{LT-mapper} supports two types of map management: \textit{live map} and \textit{meta map}. In the live map, the up-to-date representation of a scene is efficiently maintained. In the meta map, non-volume-maximizing points are iteratively removed (red boxes) while other persistent structures remain and be enhanced.
  }
  \label{fig:metamap}


\end{figure}



\begin{figure*}[!t]
  \centering
  \captionsetup{font=footnotesize}


  \includegraphics[width=0.98\textwidth, trim = 0 0 0 0, clip]{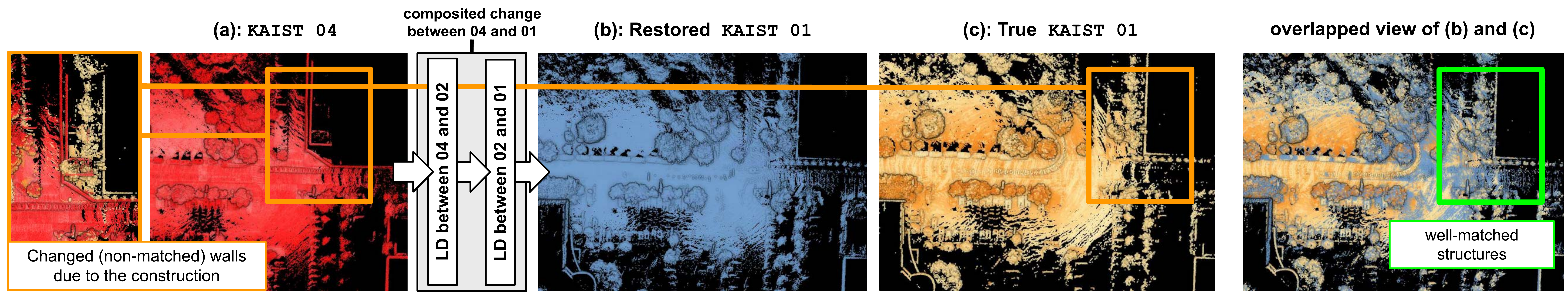}
  \caption
  {
    An example of change composition. By chaining \textit{delta maps}, \text{LT-map} can restore the map at any timestamp. After the rollback with change compositions, the unmatched wall in \texttt{KAIST 04} (red) correctly disappears and the original wall in \texttt{KAIST 01} is successfully recovered.
  }
  \label{fig:compose}

  \vspace{-3mm}

\end{figure*}



\subsection{Lifelong Mapping}
\label{sec:explifelong}



\text{LT-mapper} can update the world representation in two ways, as shown in \figref{fig:metamap}. First, \text{LT-mapper} can efficiently maintain a \textit{live map} via sending only LD changes to a central server, instead of whole snapshots. For the second representation, \textit{meta map}, \text{LT-mapper} extends spatial volumes without adding weak PDs. This elaborates a meta representation of a 3D scene, which is independent of short-term stationary or periodic changes. 




\subsection{Change Composition}
\label{sec:compose}

Because there exist no point-wise ground-truth for the 3D changes over time, we propose an implicit way to qualitatively evaluate our LD change detection performance via composing changes (\figref{fig:compose}). We have pre-calculated LD changes from \text{LT-removert} between \texttt{KAIST 01} and \texttt{02}, \texttt{02} and \texttt{04}, except for \texttt{01} and \texttt{04}. If the calculated $\text{LD}_{\texttt{01}}^{\texttt{02}}$ and $\text{LD}_{\texttt{02}}^{\texttt{04}}$ are reliable, then the composed virtual LD change $\hat{\text{LD}}_{\texttt{01}}^{\texttt{04}}$ should match the actually obtained ${\text{LD}}_{\texttt{01}}^{\texttt{04}}$. In other words, $\mathcal{P}_{\texttt{04}} \odot {\hat{\text{LD}}}_{\texttt{04}}^{\texttt{01}}$ should be equal to $\mathcal{P}_{\texttt{01}}$. As can be seen in \figref{fig:compose}, the restored \texttt{KAIST 01} from \texttt{04} is well-matched to the real map of \texttt{KAIST 01}. This \textit{delta map chaining} process is identical to map \textit{rollback}, and we can restore a map at any timestamp without saving all memory-consuming snapshots.



\begin{table}[t!]

    \vspace{-1mm}
    \centering 
    \captionsetup{font=scriptsize}

    {\scriptsize 

    \resizebox{\columnwidth}{!}{
      \begin{tabular}{c | ccc | cccc} 
      \toprule[1.0pt] 
        Chamfer Distance                                                             & \multirow{2}{*}{Max}   & \multirow{2}{*}{Avg}  & \multirow{2}{*}{Var}   & \multicolumn{3}{c}{$NP_{CD>\tau}$} & \multirow{2}{*}{$NP_{valid}$}\\
        (CD)                                                                         &       &      &       & $\tau=1$ & $\tau=2$ & $\tau=3$ & \\[-.1ex]
      \midrule[1.0pt]\\[-3.ex]
        \makecell{Pos. Pair (\texttt{01} $\leftrightarrow$ \texttt{Restored 01})}  & 6.41  & 0.29 & 0.31  & 38       & 9        & 1        & 1424 \\[-.3ex]
        \midrule[.3pt]\\[-3.ex]
        \makecell{Neg. Pair (\texttt{01} $\leftrightarrow$ \texttt{04})}           & 29.22 & 0.51 & 1.67  & 100      & 20       & 8        & 1386 \\[-.1ex]
      \bottomrule[1.0pt]
      \end{tabular}
    }
    }

    \caption
    {
        Accuracy evaluation of \figref{fig:compose} using Chamfer distance and its statistics to summarize the structural inconsistency. Among the number of patches (NP) containing at least 25 points ($NP_{valid}$), we count NP having the distance value larger than a threshold ($\tau$).
    } 

    \label{tab:composeacc}
    \vspace{-2mm}
\end{table}




\begin{table}[t!]
    \captionsetup{font=scriptsize}

    \vspace{-0mm}


    \centering 

    {\scriptsize 

    \resizebox{0.99\columnwidth}{!}{
    \begin{tabular}{c | c c} 
    \toprule[1.0pt]

    \makecell{NOTE: for 100 $\text{KF}_{\texttt{01}}$s \\and near 200 $\text{KF}_{\texttt{04}}$s in \figref{fig:compose}}       & \makecell{Memory Usage [MB]\\ (merged map)} & \makecell{Computation Time [sec]\\(between \texttt{04} and \texttt{01})} \\[.1ex]

    \midrule[1.0pt]\\[-3.ex]
    \makecell{Baseline (saving whole snapshot)}  & 213.6 & \makecell{87.0 (w/o HD removal)\\160.0 (w/ HD removal)}  \\[-.4ex]
    \midrule[.3pt]\\[-3.ex]
    \makecell{Ours (LT-map, delta map chaining)} &       85.7                        & 9.8    \\[0.ex]
    \hline\\[-1.6ex]
    Efficiency ratio                  &         60\% saved                      & 8.9 (16.3) times faster  \\

    \bottomrule

    \end{tabular}
    }
    }


    \caption
    {
       Efficiency evaluation of \figref{fig:compose}. \text{LT-mapper} efficiency against the snapshot-based method.
    } 

    \label{tab:compose}

    \vspace{-5mm}
\end{table}



\noindent \textbf{Accuracy:} To quantitatively evaluate the consistency between the \texttt{restored 01} and \texttt{01}, we use Chamfer distance \cite{fan2017point}. First, we divide the aligned maps in \figref{fig:compose} into \unit{5}{m}${}^{3}$ cubic patches and calculate the distance for each pair of corresponding patches having more than at least 25 points in a cubic. \tabref{tab:composeacc} shows the positive pair (i.e., \texttt{restored 01} and \texttt{01}) had lower distances and less inconsistent patches for a given target map \texttt{01} than the negative pair (e.g., \texttt{04} and \texttt{01}) are reported.


\noindent \textbf{Efficiency:} For spatial change analysis, our change composition has also an advantage in computational efficiency. The time cost of conducting \text{LT-removert} once is $\mathcal{O}(nm)$, where $n$ is the number of keyframes and $m$ is the number of map points. Running \text{LT-removert} for a pair of consecutive sessions (i.e., between time $t$ and $t+1$) is required only once. Later when we aim to compare two arbitrary sessions, we need only to perform the lightweight change composition (empirically \unit{0.05}{sec} per keyframe) which is linear to the $n$. Only from consecutive changes, we can efficiently make any combinatorial pair of sessions' changes without re-performing \text{LT-removert} for the pair. The quantitative report is summarized in \tabref{tab:compose}. Compared to maintaining entire sessions, our \text{LT-map} representation with \textit{delta map} saved nearly \unit{60}{\%} of the amount of memory and yielded a performance $8.9$ times faster than computing LD changes from scratch.











\begin{figure}[!t]
  \centering
  \captionsetup{font=footnotesize}

  \includegraphics[width=0.85\columnwidth, trim = 0 0 0 0, clip]{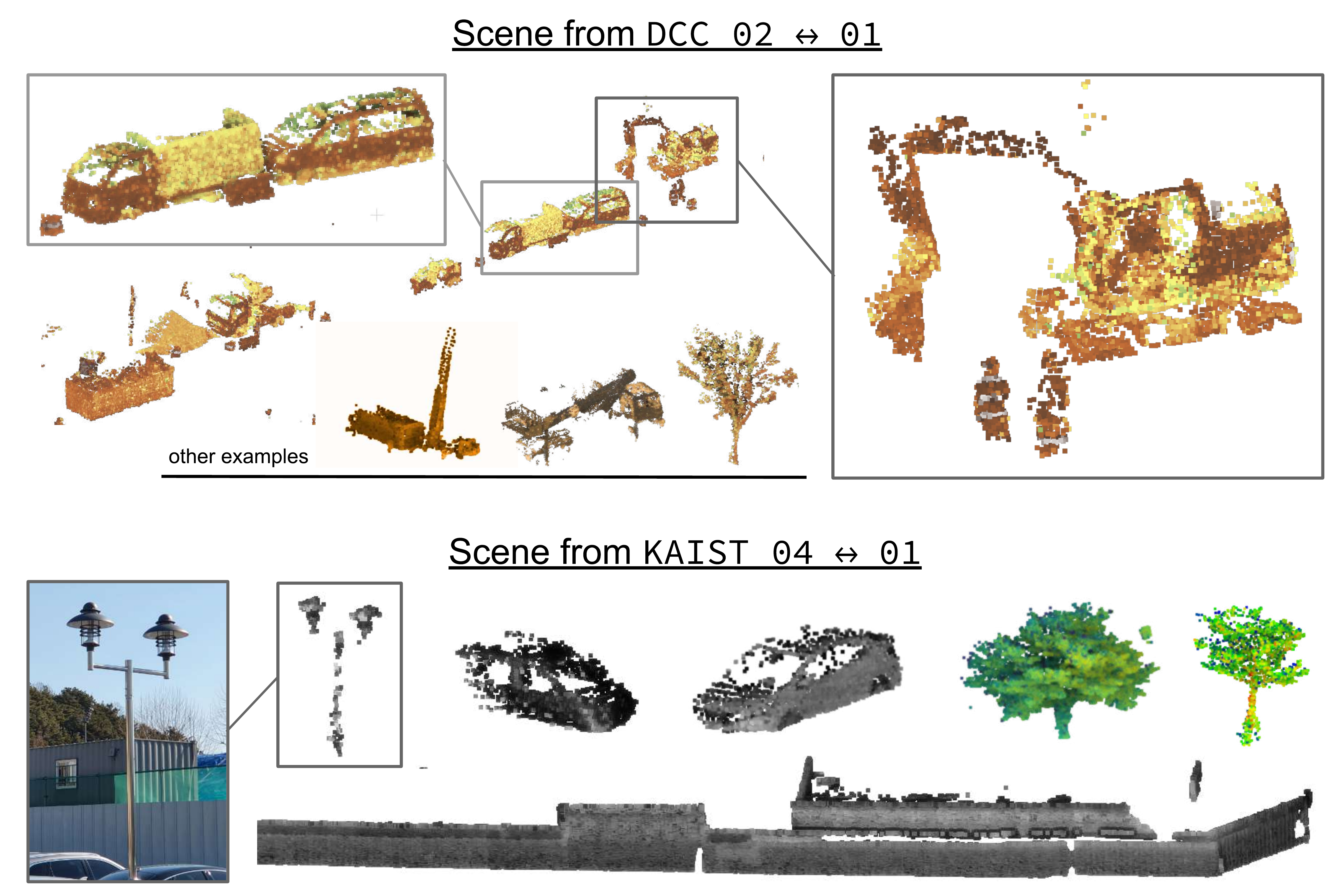}
  \caption
  {
      Automatically parsed object examples. Top: the strong ND points in the scene of \figref{fig:ldcases}. Bottom: examples from the scene of \figref{fig:cover}. Colors were arbitrarily selected for clear visualization. All objects are parsed with simple Euclidean distance-based clustering.
  }
  \label{fig:obj}

  \vspace{-3mm}

\end{figure}


\subsection{Automatic Parsing of Ephemeral Objects}
\label{sec:obj}

From our PD and ND maps, we can easily segment ephemeral objects' points, as in \figref{fig:obj}. We expect this to promote understanding of the relationship between the ephemerality of an object and its 3D shapes. If we proactively assess the ephemerality of a 3D object, we also expect this to improve \text{LT-removert} performance via serving it as prior information.


\section{Conclusion}
\label{sec:conclusion}
\vspace{-1mm}

In this work, we presented an open, modular, and unified LiDAR-based lifelong mapping framework, \textbf{LT-mapper}. We tackled three challenges to build a reliable long-term (day to year scale) map update system: \textit{1}) no or inconsistent ground-truths over sessions, \textit{2}) noisy points from high dynamic objects, and \textit{3}) new\text{/}disappeared structures. As shown in our extensive evaluations of real-world changing urban environments, our open framework can be a core engine for multiple applications for urban spatial understanding by efficiently maintaining \text{live/meta maps}, composing changes, and automatically sorting ephemeral shapes. 

\clearpage

\renewcommand*{\bibfont}{\small}
\bibliographystyle{IEEEtranN} 
\bibliography{string-short,references}

\begin{thebibliography}{39}
\providecommand{\natexlab}[1]{#1}
\providecommand{\url}[1]{#1}
\csname url@samestyle\endcsname
\providecommand{\newblock}{\relax}
\providecommand{\bibinfo}[2]{#2}
\providecommand{\BIBentrySTDinterwordspacing}{\spaceskip=0pt\relax}
\providecommand{\BIBentryALTinterwordstretchfactor}{4}
\providecommand{\BIBentryALTinterwordspacing}{\spaceskip=\fontdimen2\font plus
\BIBentryALTinterwordstretchfactor\fontdimen3\font minus
  \fontdimen4\font\relax}
\providecommand{\BIBforeignlanguage}[2]{{%
\expandafter\ifx\csname l@#1\endcsname\relax
\typeout{** WARNING: IEEEtranN.bst: No hyphenation pattern has been}%
\typeout{** loaded for the language `#1'. Using the pattern for}%
\typeout{** the default language instead.}%
\else
\language=\csname l@#1\endcsname
\fi
#2}}
\providecommand{\BIBdecl}{\relax}
\BIBdecl

\bibitem[Pomerleau et~al.(2014)Pomerleau, Kr{\"u}si, Colas, Furgale, and
  Siegwart]{pomerleau2014long}
F.~Pomerleau, P.~Kr{\"u}si, F.~Colas, P.~Furgale, and R.~Siegwart, ``Long-term
  3d map maintenance in dynamic environments,'' in \emph{Proc. {IEEE} Intl.
  Conf. on Robot. and Automat.}\hskip 1em plus 0.5em minus 0.4em\relax IEEE,
  2014, pp. 3712--3719.

\bibitem[Kim et~al.(2010)Kim, Kaess, Fletcher, Leonard, Bachrach, Roy, and
  Teller]{kim2010multiple}
B.~Kim, M.~Kaess, L.~Fletcher, J.~Leonard, A.~Bachrach, N.~Roy, and S.~Teller,
  ``Multiple relative pose graphs for robust cooperative mapping,'' in
  \emph{Proc. {IEEE} Intl. Conf. on Robot. and Automat.}, 2010, pp. 3185--3192.

\bibitem[Ambru{\c{s}} et~al.(2014)Ambru{\c{s}}, Bore, Folkesson, and
  Jensfelt]{ambrucs2014meta}
R.~Ambru{\c{s}}, N.~Bore, J.~Folkesson, and P.~Jensfelt, ``Meta-rooms: Building
  and maintaining long term spatial models in a dynamic world,'' in \emph{Proc.
  {IEEE}/{RSJ} Intl. Conf. on Intell. Robots and Sys.}\hskip 1em plus 0.5em
  minus 0.4em\relax IEEE, 2014, pp. 1854--1861.

\bibitem[Kim et~al.(2020)Kim, Park, Cho, Jeong, and Kim]{kim2020mulran}
G.~Kim, Y.~S. Park, Y.~Cho, J.~Jeong, and A.~Kim, ``{MulRan: Multimodal Range
  Dataset for Urban Place Recognition},'' in \emph{Proc. {IEEE} Intl. Conf. on
  Robot. and Automat.}, 2020.

\bibitem[Walcott-Bryant et~al.(2012)Walcott-Bryant, Kaess, Johannsson, and
  Leonard]{walcott2012dynamic}
A.~Walcott-Bryant, M.~Kaess, H.~Johannsson, and J.~J. Leonard, ``Dynamic pose
  graph slam: Long-term mapping in low dynamic environments,'' in \emph{Proc.
  {IEEE}/{RSJ} Intl. Conf. on Intell. Robots and Sys.}\hskip 1em plus 0.5em
  minus 0.4em\relax IEEE, 2012, pp. 1871--1878.

\bibitem[Wellhausen et~al.(2017)Wellhausen, Dub{\'e}, Gawel, Siegwart, and
  Cadena]{wellhausen2017reliable}
L.~Wellhausen, R.~Dub{\'e}, A.~Gawel, R.~Siegwart, and C.~Cadena, ``Reliable
  real-time change detection and mapping for 3d lidars,'' in \emph{2017 IEEE
  International Symposium on Safety, Security and Rescue Robotics
  (SSRR)}.\hskip 1em plus 0.5em minus 0.4em\relax IEEE, 2017, pp. 81--87.

\bibitem[Schauer and N{\"u}chter(2018)]{schauer2018peopleremover}
J.~Schauer and A.~N{\"u}chter, ``{The Peopleremover — Removing Dynamic
  Objects From 3-D Point Cloud Data by Traversing a Voxel Occupancy Grid},''
  \emph{{IEEE} Robot. and Automat. Lett.}, vol.~3, no.~3, pp. 1679--1686, 2018.

\bibitem[Kim and Kim(2020)]{gskim-2020-iros}
G.~Kim and A.~Kim, ``{Remove, then Revert: Static Point cloud Map Construction
  using Multiresolution Range Images },'' in \emph{Proc. {IEEE}/{RSJ} Intl.
  Conf. on Intell. Robots and Sys.}, Las Vegas, Oct. 2020.

\bibitem[Kim et~al.(2018)Kim, Jeong, and Kim]{kim2018stereo}
Y.~Kim, J.~Jeong, and A.~Kim, ``Stereo camera localization in 3d lidar maps,''
  in \emph{Proc. {IEEE}/{RSJ} Intl. Conf. on Intell. Robots and Sys.}, 2018,
  pp. 1--9.

\bibitem[Kim and Kim(2018)]{kim2018scan}
G.~Kim and A.~Kim, ``{Scan Context: Egocentric spatial descriptor for place
  recognition within 3D point cloud map},'' in \emph{Proc. {IEEE}/{RSJ} Intl.
  Conf. on Intell. Robots and Sys.}, 2018, pp. 4802--4809.

\bibitem[Alcantarilla et~al.(2016)Alcantarilla, Stent, Ros, Arroyo, and
  Gherardi]{Stent-RSS-16}
P.~F. Alcantarilla, S.~Stent, G.~Ros, R.~Arroyo, and R.~Gherardi,
  ``{Street-View Change Detection with Deconvolutional Networks},'' in
  \emph{Proceedings of Robotics: Science and Systems}, AnnArbor, Michigan, June
  2016.

\bibitem[Labb{\'e} and Michaud(2019)]{labbe2019rtab}
M.~Labb{\'e} and F.~Michaud, ``Rtab-map as an open-source lidar and visual
  simultaneous localization and mapping library for large-scale and long-term
  online operation,'' \emph{Journal of Field Robotics}, vol.~36, no.~2, pp.
  416--446, 2019.

\bibitem[Schneider et~al.(2018)Schneider, Dymczyk, Fehr, Egger, Lynen,
  Gilitschenski, and Siegwart]{schneider2018maplab}
T.~Schneider, M.~Dymczyk, M.~Fehr, K.~Egger, S.~Lynen, I.~Gilitschenski, and
  R.~Siegwart, ``maplab: An open framework for research in visual-inertial
  mapping and localization,'' \emph{{IEEE} Robot. and Automat. Lett.}, vol.~3,
  no.~3, pp. 1418--1425, 2018.

\bibitem[Elvira et~al.(2019)Elvira, Tard{\'{o}}s, and Montiel]{orbatlas19}
R.~Elvira, J.~D. Tard{\'{o}}s, and J.~M.~M. Montiel, ``Orbslam-atlas: a robust
  and accurate multi-map system,'' in \emph{Proc. {IEEE}/{RSJ} Intl. Conf. on
  Intell. Robots and Sys.}, 2019, pp. 6253--6259.

\bibitem[Campos et~al.(2020)Campos, Elvira, G\'omez, Montiel, and
  Tard\'os]{orbslam3}
C.~Campos, R.~Elvira, J.~J. G\'omez, J.~M.~M. Montiel, and J.~D. Tard\'os,
  ``{ORB-SLAM3: An Accurate Open-Source Library for Visual, Visual-Inertial and
  Multi-Map SLAM},'' \emph{arXiv preprint arXiv:2007.11898}, 2020.

\bibitem[Palazzolo and Stachniss(2018)]{palazzolo2018fast}
E.~Palazzolo and C.~Stachniss, ``Fast image-based geometric change detection
  given a 3d model,'' in \emph{Proc. {IEEE} Intl. Conf. on Robot. and
  Automat.}\hskip 1em plus 0.5em minus 0.4em\relax IEEE, 2018, pp. 6308--6315.

\bibitem[Tipaldi et~al.(2013)Tipaldi, Meyer-Delius, and
  Burgard]{tipaldi2013lifelong}
G.~D. Tipaldi, D.~Meyer-Delius, and W.~Burgard, ``Lifelong localization in
  changing environments,'' \emph{The International Journal of Robotics
  Research}, vol.~32, no.~14, pp. 1662--1678, 2013.

\bibitem[Sun et~al.(2018)Sun, Yan, Zaganidis, Zhao, and
  Duckett]{sun2018recurrent}
L.~Sun, Z.~Yan, A.~Zaganidis, C.~Zhao, and T.~Duckett, ``Recurrent-octomap:
  Learning state-based map refinement for long-term semantic mapping with
  3-d-lidar data,'' \emph{IEEE Robotics and Automation Letters}, vol.~3, no.~4,
  pp. 3749--3756, 2018.

\bibitem[Banerjee et~al.(2019)Banerjee, Lisin, Briggs, Llofriu, and
  Munich]{banerjee2019lifelong}
N.~Banerjee, D.~Lisin, J.~Briggs, M.~Llofriu, and M.~E. Munich, ``{Lifelong
  Mapping using Adaptive Local Maps},'' in \emph{2019 European Conference on
  Mobile Robots (ECMR)}.\hskip 1em plus 0.5em minus 0.4em\relax IEEE, 2019, pp.
  1--8.

\bibitem[Krajn{\'\i}k et~al.(2017)Krajn{\'\i}k, Fentanes, Santos, and
  Duckett]{krajnik2017fremen}
T.~Krajn{\'\i}k, J.~P. Fentanes, J.~M. Santos, and T.~Duckett, ``Fremen:
  Frequency map enhancement for long-term mobile robot autonomy in changing
  environments,'' \emph{IEEE Transactions on Robotics}, vol.~33, no.~4, pp.
  964--977, 2017.

\bibitem[Ding et~al.(2020)Ding, Hou, Gao, Wan, and Song]{ding2020lidar}
W.~Ding, S.~Hou, H.~Gao, G.~Wan, and S.~Song, ``Lidar inertial odometry aided
  robust lidar localization system in changing city scenes,'' in \emph{Proc.
  {IEEE} Intl. Conf. on Robot. and Automat.}\hskip 1em plus 0.5em minus
  0.4em\relax IEEE, 2020, pp. 4322--4328.

\bibitem[Shan and Englot(2018)]{shan2018lego}
T.~Shan and B.~Englot, ``{LeGO-LOAM: Lightweight and ground-optimized lidar
  odometry and mapping on variable terrain},'' in \emph{Proc. {IEEE}/{RSJ}
  Intl. Conf. on Intell. Robots and Sys.}, 2018, pp. 4758--4765.

\bibitem[Cho et~al.(2020)Cho, Kim, and Kim]{cho2020unsupervised}
Y.~Cho, G.~Kim, and A.~Kim, ``Unsupervised geometry-aware deep lidar
  odometry,'' in \emph{Proc. {IEEE} Intl. Conf. on Robot. and Automat.}\hskip
  1em plus 0.5em minus 0.4em\relax IEEE, 2020, pp. 2145--2152.

\bibitem[Li and Wang(2020)]{li2020dmlo}
Z.~Li and N.~Wang, ``{DMLO: Deep Matching LiDAR Odometry},'' 2020.

\bibitem[Shan et~al.(2020)Shan, Englot, Meyers, Wang, Ratti, and
  Rus]{shan2020lio}
T.~Shan, B.~Englot, D.~Meyers, W.~Wang, C.~Ratti, and D.~Rus, ``Lio-sam:
  Tightly-coupled lidar inertial odometry via smoothing and mapping,'' in
  \emph{Proc. {IEEE}/{RSJ} Intl. Conf. on Intell. Robots and Sys.}, 2020.

\bibitem[Yokozuka et~al.(2020)Yokozuka, Koide, Oishi, and
  Banno]{yokozukalitamin}
M.~Yokozuka, K.~Koide, S.~Oishi, and A.~Banno, ``Litamin: Lidar-based tracking
  and mapping by stabilized icp for geometry approximation with normal
  distributions,'' 2020.

\bibitem[He et~al.(2016)He, Wang, and Zhang]{he2016m2dp}
L.~He, X.~Wang, and H.~Zhang, ``\text{M2DP:} a novel \text{3D} point cloud
  descriptor and its application in loop closure detection,'' in \emph{Proc.
  {IEEE}/{RSJ} Intl. Conf. on Intell. Robots and Sys.}, 2016, pp. 231--237.

\bibitem[Uy and Lee(2018)]{angelina2018pointnetvlad}
M.~A. Uy and G.~H. Lee, ``{PointNetVLAD: Deep point cloud based retrieval for
  large-scale place recognition},'' in \emph{Proc. {IEEE} Conf. on Comput.
  Vision and Pattern Recog.}, 2018, pp. 4470--4479.

\bibitem[Chen et~al.(2019)Chen, L{\"a}be, Milioto, R{\"o}hling, Vysotska, Haag,
  Behley, and Stachniss]{Chen2019OverlapNetLC}
X.~Chen, T.~L{\"a}be, A.~Milioto, T.~R{\"o}hling, O.~Vysotska, A.~Haag,
  J.~Behley, and C.~Stachniss, ``{OverlapNet: Loop Closing for LiDAR-based
  SLAM},'' in \emph{Proc. Robot.: Science \& Sys. Conf.}, 2019.

\bibitem[Xu et~al.(2020)Xu, Yin, Chen, Wang, and Xiong]{xu2020disco}
X.~Xu, H.~Yin, Z.~Chen, Y.~Wang, and R.~Xiong, ``{DiSCO: Differentiable Scan
  Context with Orientation},'' \emph{arXiv preprint arXiv:2010.10949}, 2020.

\bibitem[McDonald et~al.(2013)McDonald, Kaess, Cadena, Neira, and
  Leonard]{mcdonald2013real}
J.~McDonald, M.~Kaess, C.~Cadena, J.~Neira, and J.~J. Leonard, ``Real-time
  6-dof multi-session visual slam over large-scale environments,''
  \emph{Robotics and Autonomous Systems}, vol.~61, no.~10, pp. 1144--1158,
  2013.

\bibitem[Ozog et~al.(2016)Ozog, Carlevaris-Bianco, Kim, and
  Eustice]{ozog2016long}
P.~Ozog, N.~Carlevaris-Bianco, A.~Kim, and R.~M. Eustice, ``Long-term mapping
  techniques for ship hull inspection and surveillance using an autonomous
  underwater vehicle,'' \emph{Journal of Field Robotics}, vol.~33, no.~3, pp.
  265--289, 2016.

\bibitem[Smith et~al.(1990)Smith, Self, and Cheeseman]{smith1990estimating}
R.~Smith, M.~Self, and P.~Cheeseman, ``Estimating uncertain spatial
  relationships in robotics,'' in \emph{Autonomous robot vehicles}.\hskip 1em
  plus 0.5em minus 0.4em\relax Springer, 1990, pp. 167--193.

\bibitem[Agarwal et~al.(2013)Agarwal, Tipaldi, Spinello, Stachniss, and
  Burgard]{agarwal2013robust}
P.~Agarwal, G.~D. Tipaldi, L.~Spinello, C.~Stachniss, and W.~Burgard, ``{Robust
  Map Optimization using Dynamic Covariance Scaling},'' in \emph{Proc. {IEEE}
  Intl. Conf. on Robot. and Automat.}, 2013, pp. 62--69.

\bibitem[Mangelson et~al.(2018)Mangelson, Dominic, Eustice, and
  Vasudevan]{mangelson2018pairwise}
J.~G. Mangelson, D.~Dominic, R.~M. Eustice, and R.~Vasudevan, ``Pairwise
  consistent measurement set maximization for robust multi-robot map merging,''
  in \emph{Proc. {IEEE} Intl. Conf. on Robot. and Automat.}\hskip 1em plus
  0.5em minus 0.4em\relax IEEE, 2018, pp. 2916--2923.

\bibitem[Kaess et~al.(2012)Kaess, Johannsson, Roberts, Ila, Leonard, and
  Dellaert]{kaess2012isam2}
M.~Kaess, H.~Johannsson, R.~Roberts, V.~Ila, J.~J. Leonard, and F.~Dellaert,
  ``{iSAM2: Incremental smoothing and mapping using the Bayes tree},''
  \emph{The International Journal of Robotics Research}, vol.~31, no.~2, pp.
  216--235, 2012.

\bibitem[Dellaert(2012)]{dellaert2012factor}
F.~Dellaert, ``{Factor graphs and GTSAM: A hands-on introduction},'' Georgia
  Institute of Technology, Tech. Rep., 2012.

\bibitem[Zhang and Scaramuzza(2018)]{Zhang18iros}
Z.~Zhang and D.~Scaramuzza, ``{A Tutorial on Quantitative Trajectory Evaluation
  for Visual(-Inertial) Odometry},'' in \emph{Proc. {IEEE}/{RSJ} Intl. Conf. on
  Intell. Robots and Sys.}, 2018.

\bibitem[Fan et~al.(2017)Fan, Su, and Guibas]{fan2017point}
H.~Fan, H.~Su, and L.~J. Guibas, ``A point set generation network for 3d object
  reconstruction from a single image,'' in \emph{Proceedings of the IEEE
  conference on computer vision and pattern recognition}, 2017, pp. 605--613.

\end{thebibliography}


\vfill

\end{document}